\documentclass[runningheads]{llncs}
\usepackage[T1]{fontenc}
\usepackage{graphicx}
\usepackage{cite}
\usepackage{amsmath,amssymb,amsfonts}
\usepackage{algorithmic}
\usepackage{graphicx}
\usepackage{textcomp}
\usepackage{xcolor}
\usepackage{bibentry}
\usepackage{threeparttable} 
\usepackage{multirow}
\usepackage{wrapfig}
\usepackage{bbding} 
\usepackage{bbm}
\usepackage{array}
\usepackage{float}
\usepackage{booktabs}    
\usepackage{tikz}
\usepackage{comment}
\usepackage{amsmath,amssymb} 
\usepackage{subcaption}
\usepackage{url}
\usepackage[hidelinks]{hyperref}
\usepackage[T1]{fontenc}
\usepackage[utf8]{inputenc}
\usepackage{authblk}
\begin{document}
\title{AI WALKUP: A Computer-Vision Approach to Quantifying MDS-UPDRS in Parkinson’s Disease}
\titlerunning{MICCAI 2023 Workshops}
\titlerunning{AI WALKUP}
%
\author{Xiang Xiang (\Envelope) \and Zihan Zhang \and Jing Ma \and Yao Deng}
\authorrunning{Xiang et al.}
%
\institute{Key Lab of Image Processing and Intelligent Control, Ministry of Education\\ School of Artificial Intelligence and Automation\\ Huazhong University of Science and Technology, Wuhan 430074, China\\  \email{xex@hust.edu.cn}}
\maketitle              
\begin{abstract}
Parkinson's Disease (PD) is the second most common neurodegenerative disorder. The existing assessment method for PD is usually the Movement Disorder Society - Unified Parkinson’s Disease Rating Scale (MDS-UPDRS) to assess the severity of various types of motor symptoms and disease progression. However, manual assessment suffers from high subjectivity, lack of consistency, and high cost and low efficiency of manual communication. We want to use a computer vision based solution to capture human pose images based on a camera, reconstruct and perform motion analysis using algorithms, and extract the features of the amount of motion through feature engineering. The proposed approach can be deployed on different smartphones, and the video recording and artificial intelligence analysis can be done quickly and easily through our APP.

\keywords{Parkinson's Disease \and motor assessment \and pose estimation}
\end{abstract}
\section{Introduction}
\vspace{-3mm}
Parkinson's Disease (PD) is the second most common neurodegenerative disorder and an estimated 7 to 10 million people worldwide are living with Parkinson’s disease. The incidence of Parkinson’s increases with age, but an estimated 4\% of people with PD are diagnosed before the age of 50 \cite{PD-web}. The motor or movement-related symptoms include shaking, tremors, rigidity, slow movement, difficulty with walking, and problems with balance. These symptoms are progressive over time, subsequently leading to an increase in their severity.

\begin{figure}[t!]
	\begin{center}
		\includegraphics[scale=1]{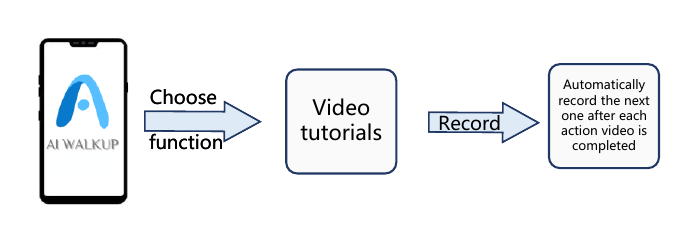}
	\end{center}
	\caption{The flowchart of the APP. The users record their specific action videos. }
	\label{fig:steps}
\end{figure}

The aim of automatic PD severity quantification is that the algorithm can predict the severity level when giving a patient’s video performing a specific task as shown in Fig.~\ref{fig:steps}. Since the MDS-UPDRS scale contains many subscales, we selected six motor function items that most affect patients' lives to aid in the assessment. Specifically, we use computer vision methods to extract patients' motion information based on the localization of human key points, thereby enabling objective and quantitative description of Parkinson's disease patients' motion status and assessment of Parkinson's disease conditions. 

The smartphone camera shooting greatly simplifies the data collection process and reduces the burden on patients and healthcare professionals. By systematically performing assessments on all patients, reduces individual subjective evaluation bias, standardizes the severity of Parkinson's disease, and assists physicians in giving relevant diagnostic and treatment opinions. It also reduces the number of offline consultations at hospitals for Parkinson's patients, reduces the cost of medical visits for patients, and improves diagnostic efficiency.

In this paper, we propose a new approach to estimate the feature of six MDS-UPDRS items including tremor, finger taps (patient taps thumb with the index finger in rapid succession), hand movements (patient opens and closes hands in rapid succession), rapid alternating movements of hands (pronation-supination movements of hands, vertically and horizontally, with as large an amplitude as possible, both hands simultaneously) and leg agility (patient taps heel on the ground in rapid succession picking up the entire leg, the amplitude should be at least 3 inches), etc. We use a computer vision-based method to extract the motion quantification features of the single monocular action videos. We do not directly predict the scores of corresponding scores of MDS-UPDRS items but the extracted features can be used to analyze the movement status of the user. The videos can be recorded using the camera of a patient's smartphone or tablet. 

\section{Related Work}

Clinically, the Movement Disorder Society - Unified Parkinson’s Disease Rating Scale (MDS-UPDRS) is a standardized tool for evaluating Parkinson's disease. MDS-UPDRS contains four parts: I for non-motor experiences of daily living, II for motor experiences of daily living, III for motor examination, and IV for motor complications \cite{martinez2013expanded}. On the scale, the motor function score accounts for more than half of the total score and it affects the normal life activities of patients most. Although MDS-UPDRS is currently the gold standard to quantify PD severity, it still has the potential to cause less reliable ratings \cite{turner2020inconsistent,evers2019measuring}. The manual rating methods have weaknesses including strong subjectivity, lack of consistency, high cost of manual communication, and low efficiency. Besides, the presence of a specialist is necessary when making the rating decisions. Thus, the manual rating is inefficient and an automatic quantification method is needed.

\section{Methodology}

\begin{figure*}[t]
	\begin{center}
		\includegraphics[scale=0.6]{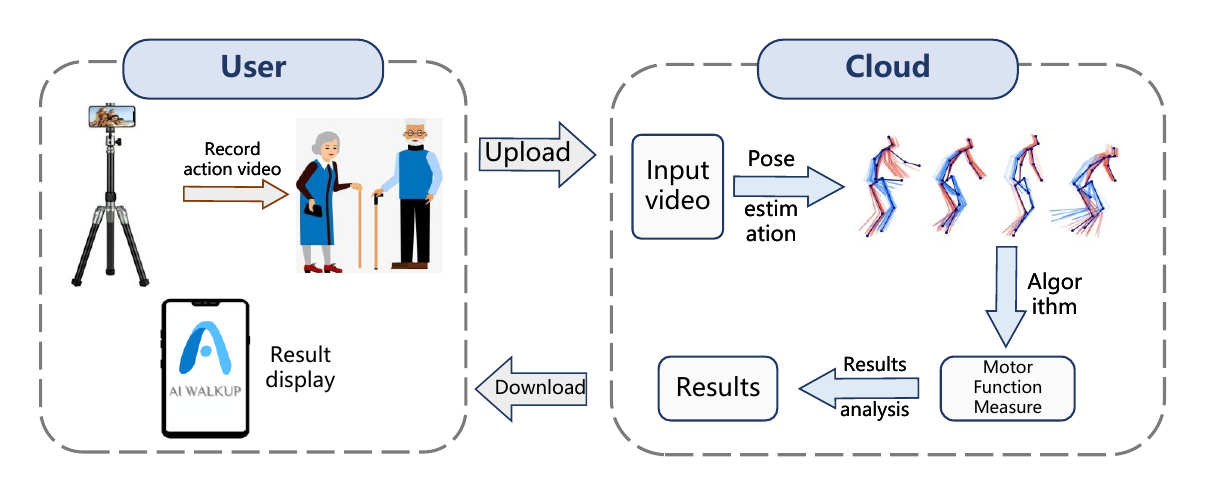}
	\end{center}
	\caption{The pipeline of our method. Our method can be divided into user side and cloud side. The user can record the action video on APP and upload the videos. The cloud can analyze the videos and return the results.}
	\label{fig:pipeline}
\end{figure*}

In this section, we introduce our method and APP implementation. The end-to-end pipeline of our method is shown in Fig.~\ref{fig:pipeline}. Our product is mainly divided into two sides: the user side, which mainly completes the local video recording and the analysis results display, and supports the recommendation of related information. After receiving the desensitized video sequences uploaded by the user, the cloud-first performs pose estimation analysis. After getting the analyzed key points of the human body, our algorithm obtains the user's movement assessment, and finally, the visualization results and analysis are linked to the user side.

\subsection{Data}
The action video data are collected using a smartphone camera. We collect the videos on 20 healthy people. For each people, we record six MDS-UPDRS items as mentioned. In further work, we can extend our methods to PD patients easily by applying the algorithm to the PD action video.

\subsection{Body keypoints}
MediaPipe Pose \cite{pose-web} is an ML solution for high-fidelity body pose tracking, inferring 33 3D landmarks, and background segmentation mask on the whole body from RGB video frames. Current pose estimation methods rely mainly on desktop environments for inference, while the MediaPipe is fast, allowing real-time inference on most smartphones.

\begin{figure}[h]
	\begin{center}
		\includegraphics[scale=0.35]{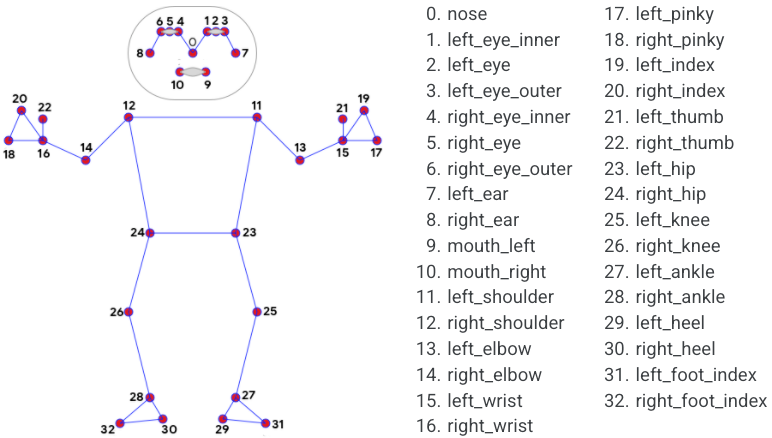}
	\end{center}
 \caption{The key points of the human body.}
	\label{fig:pose}
\end{figure}

MediaPipe Pose was chosen because of its fast and robust approach to person keypoint detection. Using a detector, Mediapipe first locates the region of interest (ROI) of the person within the frame. The tracker then uses the ROI cropped video frame as input to predict the pose landmarks. For video samples, the detector is called only when needed, i.e., when the first frame and the tracker can not recognize the body pose from the previous frame. For the other frames, Mediapipe estimates the ROI based on the information from the previous frame.

\subsection{Hand keypoints}

MediaPipe Hand \cite{hand-web} utilizes an ML pipeline consisting of multiple models working together: A palm detection model that operates on the full image and returns 3D hand key points. One hand model has 21 key points as shown in Fig.~\ref{fig:hand}.

\subsection{Signals}
\begin{figure}[h]
	\begin{center}
		\includegraphics[scale=0.15]{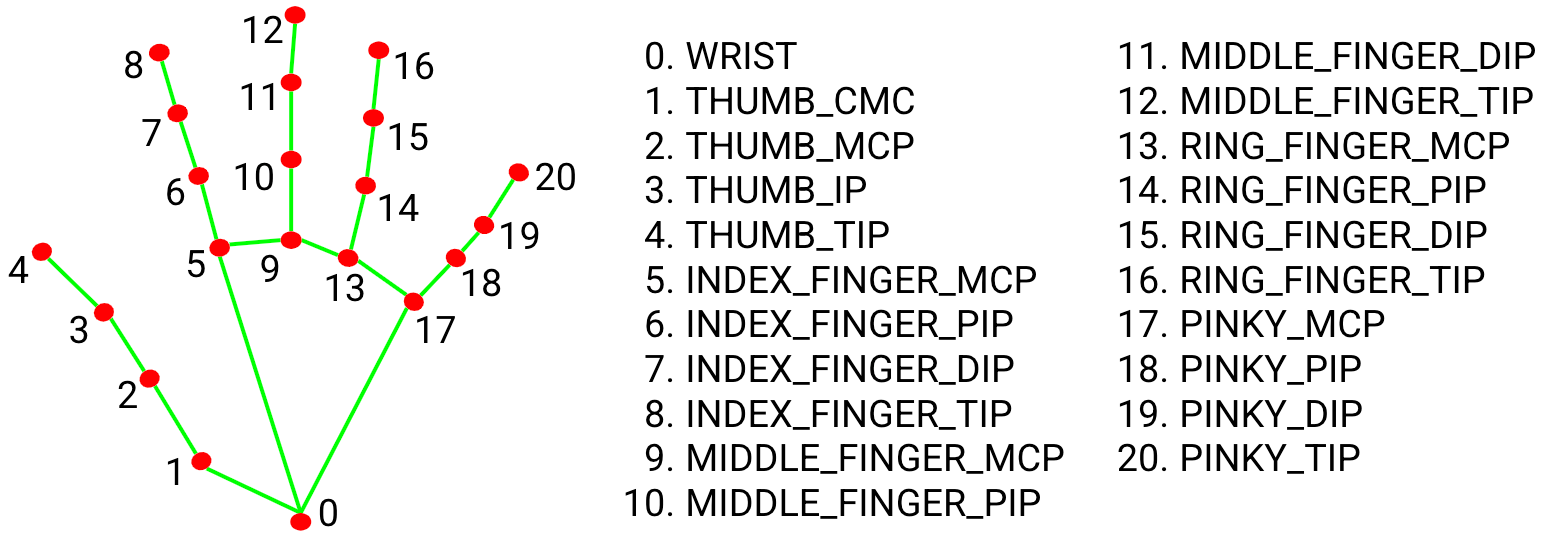}
	\end{center}
	\caption{The key points of human hands.}
	\label{fig:hand}
\end{figure}

To better estimate the motion information of the users, we designed six motion signals for each of the six action videos. We use $P$ and $H$ and the key point number in Fig.~\ref{fig:pose} and \ref{fig:hand}to indicate the body and hand points, i.e., $\boldsymbol{P_{25}}$ and $\boldsymbol{H_{0}}$ represent the left knee and wrist respectively.

\textbf{Finger taps.} Patients tap thumbs with the index finger in rapid succession. We want to check if the patient's movement is slow and/or reduced in amplitude.

\begin{equation}
    A_1 = <\boldsymbol{H_{8,0}},\boldsymbol{H_{4,0}}>
\end{equation}

The angle calculates the angle between the fingertip and thumb tip, estimating the openness of the hand. If the user taps the finger fast and with high amplitude, the signal $A_1$ changes fast and the peak is high.

\textbf{Hand movement.} The patient opens and closes hands in rapid succession. We want to check if the patient's movement is slow and/or reduced in amplitude.

\begin{equation}
    D_2 = \frac{|\boldsymbol{H_{8,0}}|+|\boldsymbol{H_{12,0}}|+|\boldsymbol{H_{16,0}}|+|\boldsymbol{H_{20,0}}|}{4}
\end{equation}

We calculate the average distance between four fingertips and the wrist. The hand size of the patient is not identical, thus we measure the change in hand distance. If the value is large, the patient opens his hand fast and with high amplitude, which means the patient is healthy.

\textbf{Rapid alternating movements of hands.} In this item, the patient performs forward flexion and backward extension movements of the hands in the vertical and horizontal directions with as much amplitude as possible, with both hands simultaneously.  

\begin{equation}
    A_3 = <\boldsymbol{H_{4,20}}, Horizontal\ Line>
\end{equation}

We calculate the angle between the hand and the horizontal line, which indicates the alternating movement angle of the hand.

\textbf{Tremor at rest.} In this item, the patient extends their arms and holds them straight for 10 seconds. We want to check if there exists tremor for the patient. Specifically, we calculate the movement of every body and hand key point. If the key points have large movement, we set $T_4=1$, else $T_4=0$.

\textbf{Leg agility.} The patient taps their heel on the ground in rapid succession picking up the entire leg. The amplitude should be at least 3 inches. It is hard for some patients to arise their legs with high amplitude and the speed is slow.

\begin{equation}
\begin{aligned}
    A_{5r} &= <\boldsymbol{H_{24,26}},\boldsymbol{H_{24,12}}>\\
    A_{5l} &= <\boldsymbol{H_{23,25}},\boldsymbol{H_{23,11}}>
\end{aligned}
\end{equation}

We calculate the angles between the leg and the upper body including the left part and right part. Some patients have difficulty to arise their left/right leg, thus we separate the two-part signals.

\textbf{Feet tap the ground.}  Patients land on their heels and lift their feet as hard as they can, lifting them quickly several times. Some patients have very inflexible foot movements and we want to assess the patient's foot movements using the following signal.

\begin{equation}
\begin{aligned}
    A_{5r} &= <\boldsymbol{H_{28,26}},\boldsymbol{H_{28,32}}>\\
    A_{5l} &= <\boldsymbol{H_{27,25}},\boldsymbol{H_{27,31}}>
\end{aligned}
\end{equation}

Similar to item five, the angles between the leg and feet reflect the movement.

\begin{table}[h]
\begin{center}
\caption{The feature extraction for the designed signals (see also tsfresh \cite{christ2018time}).}
\label{table:feature}
\begin{tabular}{ccc}
    \toprule
    Feature &Parameter& Dimension\\
	\hline  
    abs\_energy &None&1 \\
    absolute\_sum\_of\_changes &\{f\_agg: mean, maxlag: 2\}&2 \\
    agg\_autocorrelation &None&1 \\
    approximate\_entropy &\{coeff: 0, k: 10\}&2 \\
    ar\_coefficient &\{attr: pvalue\}&1 \\
    augmented\_dickey\_fuller &None&1 \\
    autocorrelation &None&1 \\
    benford\_correlation &None&1 \\
    change\_quantiles &None&1 \\
    cid\_ce &None&1 \\
    mean\_abs\_change &\{aggtype: centroid\}&1 \\
    fft\_aggregated &\{coeff: 5, attr: abs\}&2 \\
    fft\_coefficient &\{lag: 10\}&1 \\
    partial\_autocorrelation &None&1 \\
    quantile &None&1 \\
    root\_mean\_square &None&1 \\
    sample\_entropy &None&1 \\
    variance &None&1 \\
    variation\_coefficient &None&1 \\
    kurtosis &\{attr: pvalue\}&1 \\
    linear\_trend &None&1 \\
    \bottomrule
\end{tabular}
\end{center}
\end{table}

\subsection{ Feature extraction }

For the six items, we design six signals to assess the movement. To further capture key characteristics of the examination, we extract features based on the designed features. Tsfresh \cite{christ2018time} provides systematic time-series feature extraction by combining established algorithms from statistics, time-series analysis, signal processing, and nonlinear dynamics with a robust feature selection algorithm. The package can extract 100s of features from the time series. Those features describe basic characteristics of the time series such as the number of peaks, the average or maximal value, or more complex features such as the time-reversal symmetry statistic. Features are selected to fit our work as shown in Table~\ref{table:feature}.

\section{Experiment and Results}

\subsection{Visualization of inputting signals}
For six action items, we design six signals respectively. We analyze the six videos of one user and visualize the signals as follows. The x-axis represents the video frame number. The red point in the figure is the peak or bottom and the red line is the average value of the signal. For each signal, we separate it into the left part and right part to assess the movement of a different parts of the body.

\begin{figure}[h]
	\begin{center}
		\includegraphics[scale=0.37]{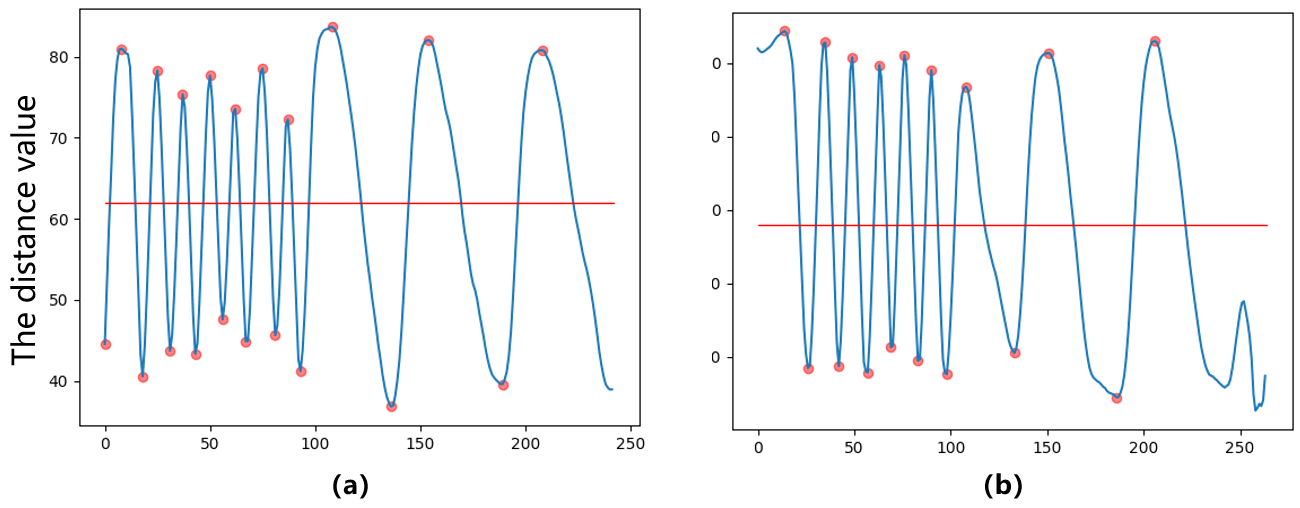}
	\end{center}
	\caption{Visualization of action two video.}
	\label{fig:action_2}
\end{figure}
\begin{figure}[h]
	\begin{center}
		\includegraphics[scale=0.4]{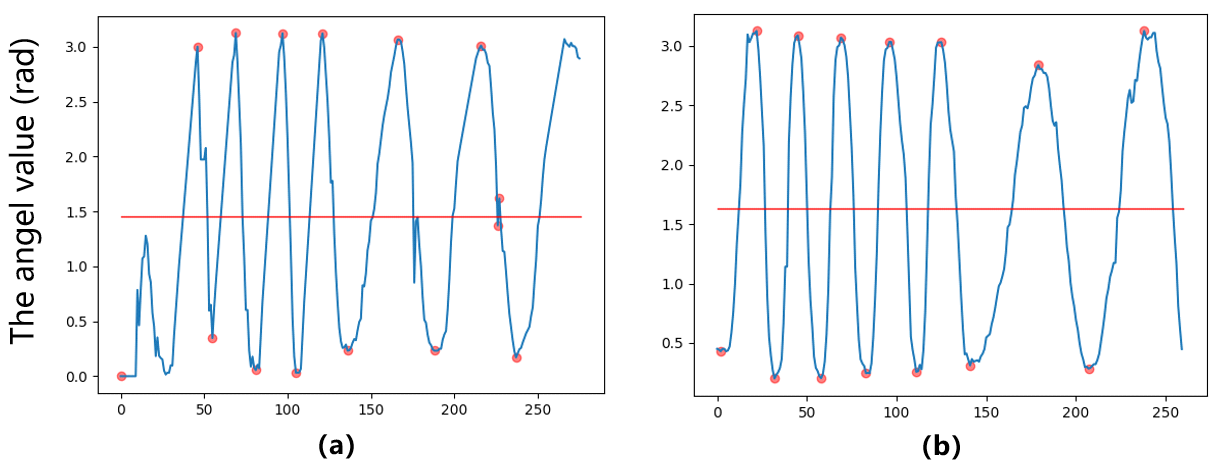}
	\end{center}
	\caption{Visualization of action three video.}
	\label{fig:action_3}
\end{figure}
\begin{figure}[h]
	\begin{center}
		\includegraphics[scale=0.35]{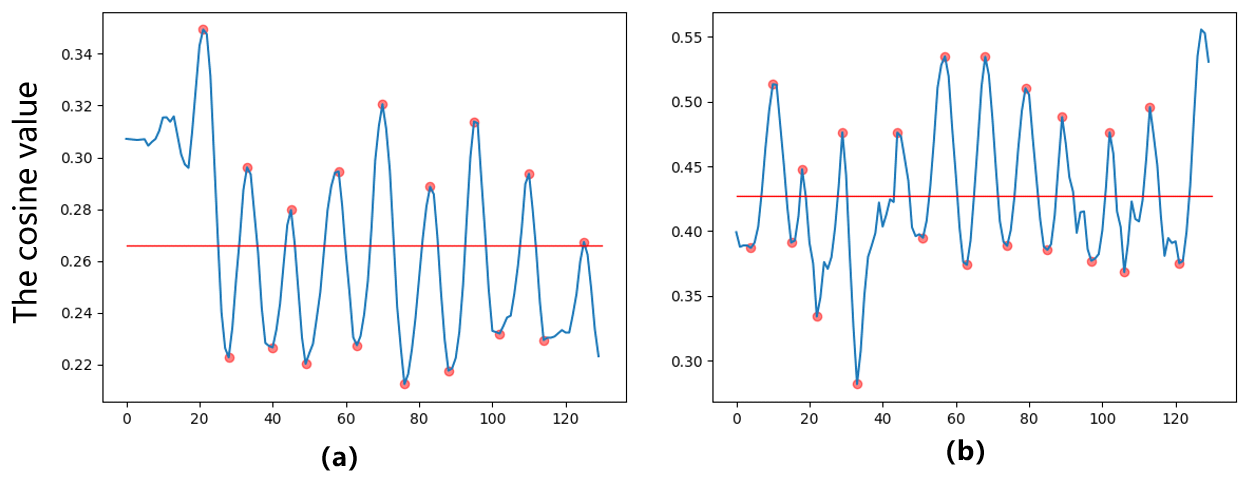}
	\end{center}
	\caption{Visualization of action five video.}
	\label{fig:action_5}
\end{figure}
\begin{figure}[h]
	\begin{center}
		\includegraphics[scale=0.35]{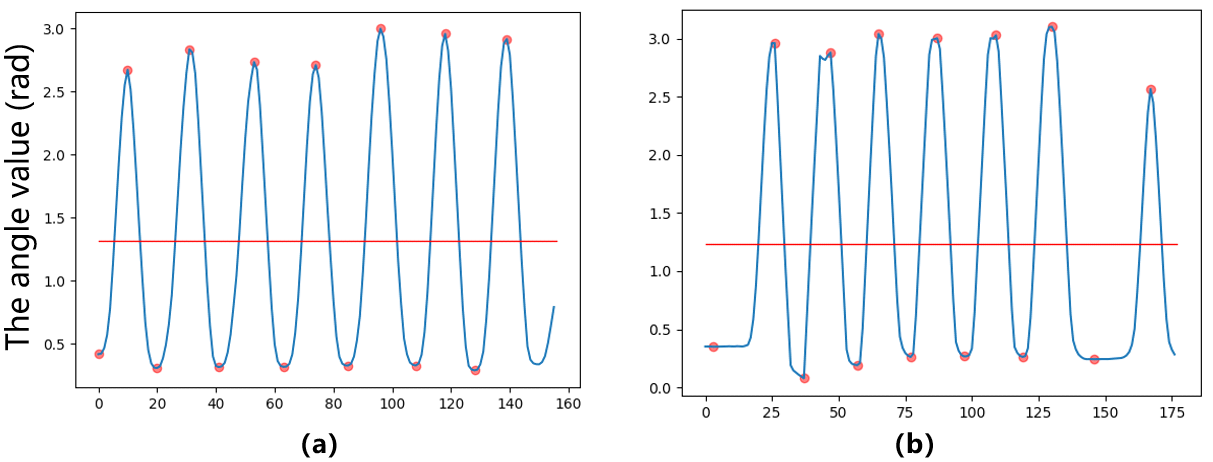}
	\end{center}
	\caption{Visualization of action six video.}
	\label{fig:action_6}
\end{figure}

Due to the limit of space, we visualize four signals. Fig.~\ref{fig:action_2} shows the time series of $D_2$, which is the average distance between four fingertips
and the wrist. We can observe that the maximum and minimum of the signal do not change much but the speed is becoming slower in the latter part. That means the user may have difficulty keeping on doing the action.

Fig.~\ref{fig:action_3} shows the time series of $A_3$. In this item, the user performs forward flexion and backward extension movements of the hands in the vertical and horizontal directions with as much amplitude as possible. The action amplitude does not change with time, but the action becomes slower in the latter stage as the time gap gets larger.
Similarly, Fig.~\ref{fig:action_5} and \ref{fig:action_6} show the corresponding signal.

\begin{figure}[h!]
	\centering
	
  \begin{minipage}[h]{0.24\textwidth}
	\centering
	\includegraphics[scale=0.51]{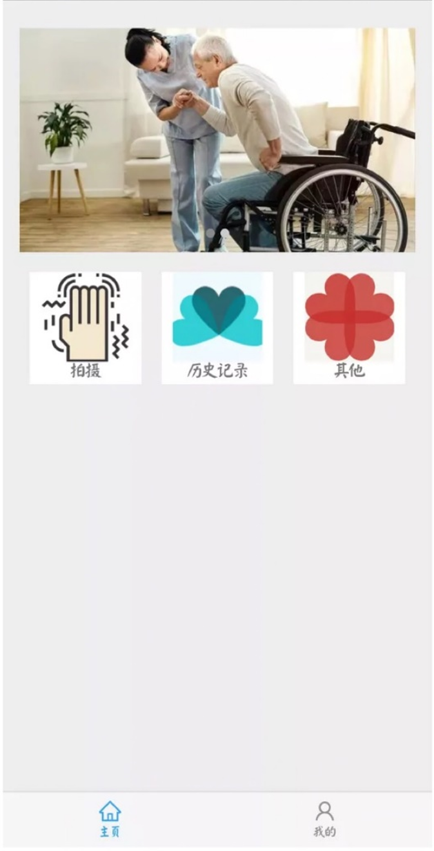}
	\label{fig:app_1}
  \end{minipage}
  \begin{minipage}[h]{0.23\textwidth}
	\centering
	\includegraphics[scale=0.45]{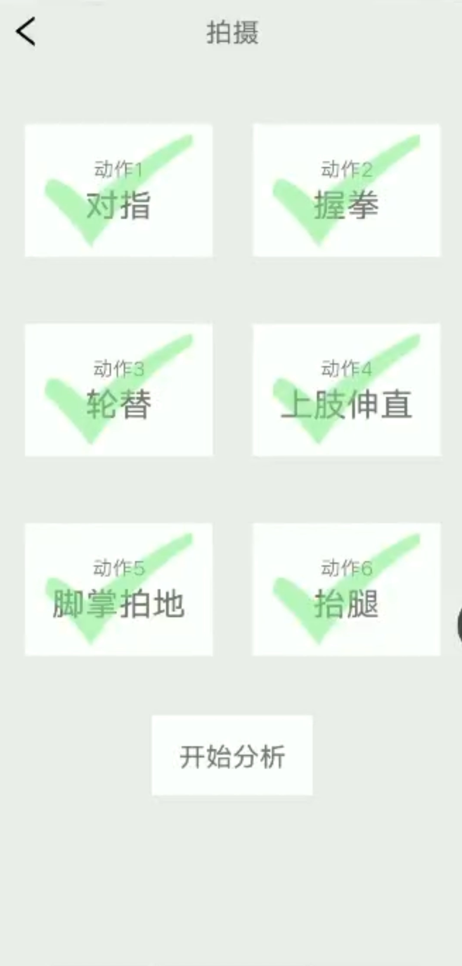}
	\label{fig:app_2}
  \end{minipage}
  \begin{minipage}[h]{0.23\textwidth}
	\centering
	\includegraphics[scale=0.45]{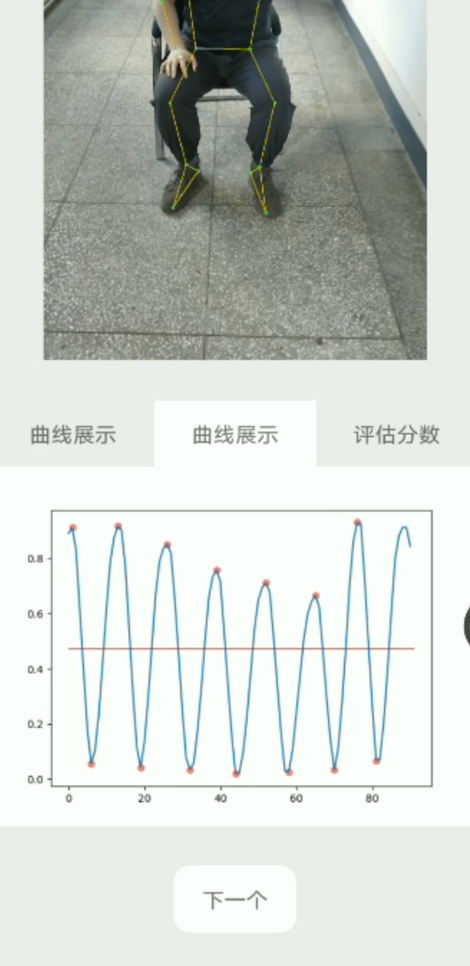}
	\label{fig:app_3}
  \end{minipage}
  	\caption{The screenshots of the APP running.}
\end{figure}

\subsection{The APP and demo results}
Our product has the corresponding functions, and the APP product is currently deployed on the Android platform. It includes a demonstration video of six actions with a shooting function. After shooting, users can enter the history module to check the analysis results of the shooting video.

The APP allows patients' family members or patients themselves to check the subjective questionnaire as well as follow the APP video instructions to record a video. The feature extraction system not only allows patients to make initial judgments about their condition but also serves as a preliminary score to assist neurologists in treating patients, effectively reducing the workload of doctors. In addition, the app can also intelligently recommend relevant information and auxiliary treatment plans to users based on their condition.

In terms of APP usage, just like the fitness APPs such as Keep, our APP has standard video demonstration guidelines that patients or potential patients' family members can follow to record scale action videos. Once users enter the APP, they can record in one click according to the video guide, while patients can also record their own videos, making the overall operation simple and intelligent. The key to an intelligent analysis algorithm is how to analyze the video after recording so that doctors should have the initial screening of the algorithm before proceeding to further diagnosis. Our algorithm can play the role of initial screening and reduce the workload of medical staff. The smartphone APP is mainly used to evaluate patients with early to mid-stage Parkinson's disease with easy-to-use and adequate guidelines. Smartphones are available in every home and can be used after installing the APP. Patients launch the APP at specific multiple time points as requested by their physicians, and the artificial intelligence condition assessment is performed by capturing video through the smartphone camera.

\section{Conclusion}
In this paper, we present a preliminary study of motor assessment approach for PD patients based on computer vision techniques. The proposed approach can be deployed on different smartphones, and the video recording and artificial intelligence analysis can be done quickly and easily through our APP.
In the future, we will perform clinic user studies, for cross-patient and over-time analysis.

\section*{Acknowledgements} 
This research was supported by HUST Independent Innovation Research Fund (Methodological Research of Automatic Pain Assessment based on Behavioral Observations, 2021XXJS096). Many thanks to Tianyao Wu, Bingyi Zhou, Zhijie Xue, Ziyang Tan, Mingyu Liu, Qian Wan, Yi Ma, Yifeng Chen, Yuting Shang, Jianchang Chen, Yunshu Cai, Yanqiu Yu, Xirong Song, Chenyu Yang, Wenbo Zheng, Changyue Fu, You Wu, and so on, for once participating in the research.

%
%
%
\bibliographystyle{splncs04}
\bibliography{egbib}

\end{document}